\documentclass[letterpaper]{article}

\usepackage{natbib,alifeconf}  
\usepackage[colorinlistoftodos]{todonotes}
\usepackage{booktabs} 
\usepackage{amssymb}
\usepackage[ruled]{algorithm2e}
\usepackage{courier}
\usepackage{subcaption}
\usepackage{url}
\usepackage{caption}

\title{Safer Reinforcement Learning through Transferable Instinct Networks}
\author{Djordje Grbic and Sebastian Risi \\
\mbox{}\\
IT University of Copenhagen, Copenhagen\\
\{djgr, sebr\}@itu.dk\\
} 

\begin{document}
\maketitle

\begin{abstract}
Random exploration is one of the main mechanisms through which reinforcement learning (RL) finds well-performing policies. 
However, it can lead to undesirable or catastrophic outcomes when 
learning online in safety-critical environments. In fact, safe learning is one of the major obstacles towards real-world agents that can learn during deployment. 
One way of ensuring that agents respect hard limitations is to explicitly configure boundaries in which they can operate. While this might work in some cases, we do not always have clear a-priori information which states and actions can lead dangerously close to hazardous states. Here, we present an approach where an additional policy can override the main policy and offer a safer alternative action. 
In our instinct-regulated RL (IR$^2$L) approach, an ``instinctual'' network is trained to recognize undesirable situations, while guarding the learning policy against entering them. 
The instinct network is pre-trained on a \emph{single task} 
where it is safe to make mistakes, and transferred to environments in which learning a new task safely is critical.
We demonstrate IR$^2$L in the OpenAI Safety gym domain, in which it receives a significantly lower number of safety violations during training than a baseline RL approach while reaching similar task performance. 

\end{abstract}
\section{Introduction}
 Deep reinforcement learning (RL) has allowed many complex tasks to be solved, from playing video games to robotics  \citep{justesen2019deep, li2017deep, silver2016mastering, mahmood2018benchmarking, gauci2018horizon, vinyals2019grandmaster}. However, developing RL agents that can learn new tasks quickly while respecting safety restrictions is an unsolved challenge  \citep{raybenchmarking,wainwright2019safelife,ortega2018building}. Most 
RL approaches rely on trial and error in order to solve tasks,
and often these trials are based on random actions. Executing random actions to learn tasks is inherently problematic, especially in the real world,  since it can cause damage to the agent and its surroundings. For example, a self-driving car cannot randomly try actions until it learns a new task because it will likely cause death and material damage.

In contrast to common RL approaches, animals in nature developed instinctual behaviors that prevent them from trying out actions that are likely dangerous to their lives. These instincts are innate behaviors provided by evolution to reduce the cost of first having to learn to avoid common  dangers. For example, human infants have a congenital fear of spiders and snakes \citep{hoehl2017itsy}, likely because the evolved instinctual fear improved our ancestors’ chances of survival. Other animals, such as rats, instinctively and without any learning avoid a specific compound found in carnivore urine \citep{ferrero2011detection}. 

In this paper, we are building on the \emph{Meta-Learned Instinctual Network} (MLIN) approach \citep{grbic2020safe}, where a policy neural network is split into two major components: a main network trained for a specific task, and a fixed pre-trained instinctual network that transfers between tasks and overrides the main policy if the agent is about to execute a dangerous action.  
%
However, meta-learning can be quite expensive since it relies on two nested learning loops: an inner task-specific loop and an outer meta-learning loop. Such high computation demands can limit the type of applications that meta-learning can be applied to.

The main insight in this paper  is that the expensive meta-learning loop in MLIN is not necessary to learn safely: we can train an  instinct network efficiently on a single task where it is acceptable to make mistakes and then combine this pre-trained instinct network with a random policy to learn another task safely. An important aspect of our \emph{instinct-regulated RL} (IR$^2$L) approach  is the balance between learning and staying safe. The instinct network cannot be too restrictive (i.e.\ blocking the policy from doing anything) and has to make sure that the policy is still able to adapt. We show that this balance can be achieved by carefully tuning the hyperparameters of the reward used to train the instinct network, which includes both hazard risk minimization and task reward.  

The results in a modification of the OpenAI Safety Gym environment \citep{ray2019benchmarking} demonstrate that an instinctual network allows an agent to learn new tasks while avoiding hazards. We also show that while a typical baseline approach that consists of pre-training a policy on a task with hazards (without an instinct network) can reduce safety violations to some extend, it performs significantly worse when compared to our IR$^2$L approach.
In the future, the idea of combining a pre-trained instinctual network with other RL methods could enable safer forms of AI across a range of different tasks.

\section{Background}
This section includes a short introduction to reinforcement learning, policy gradient methods (which we use to train our models), and AI safety.

\subsection{Reinforcement learning} 
We define an environment and the task that needs to be solved as a Markov decision process (MDP) denoted as a tuple of five elements $\mathcal{T} = \langle \mathcal{S}, \mathcal{A}, \emph{r}, \emph{P}, s_0 \rangle$; where $\mathcal{S}$ is the set of possible environment state observations, $\mathcal{A}$ is the set of actions the agent can execute, $\emph{r}$ is the numerical reward the agent receives executing an action at a certain state, $\emph{r}: \mathcal{S} \times \mathcal{A} \rightarrow \mathbb{R}$, $\emph{P}(\cdot | s, a)$ is a probability distribution of states reached by executing action $a$ in the state $s$, and $s_0(\cdot)$ is the distribution of initial states.

Often, the agent interacts with the environment in episodes, where an episode is a sequence of actions that the agent executes starting with the initial state sampled from $s_0(\cdot)$ until a terminal state $s_T$. After the agent reaches $s_T$, the environment resets and the agent is initialized in one of the $s_0$ states. A sequence of $(\langle s_0, a_0, r_0, s_1 \rangle, \langle s_1, a_1, r_1, s_2 \rangle ... \langle s_i, a_i, r_i, s_T \rangle)$ tuples is called a trajectory and it is the data used to train the policy.
The cumulative reward the agent collects in an episode is called return. Return from $(s_i, a_i)$ to $s_T$ is calculated as $R_i = r_i + \sum_{j=i+1}^{T} \gamma r_j$, where $r_i$ the reward received by executing $a_i$ at state $s_i$, $\gamma$ is the reward discount factor that is treated as a fixed hyper-parameter. The agent has to find a policy that maximizes the expected return $\mathbb{E}[R_0]$.

In online reinforcement learning the task that the agent needs to solve can change, thus requiring the agent to re-adapt to maximize the expected $j$-th task-specific return $\mathbb{E}_{\mathcal{T}_j}[R_0]$. Here we assume that the tasks differ only in the task reward $\emph{r}$ that the environment gives to the agent.

\textbf{Policy gradient methods}  \citep{Williams1992} are a family of reinforcement learning methods that optimize policy parameters applying a gradient-based optimization algorithm with respect to expected episode returns $\mathbb{E}[R_0]$. A policy is an action probability distribution $\pi_\theta(a_i|s_i)$ conditioned on the current observed state $s_i$, where $\theta$ are the policy parameters. Normally, the policy is modeled with an artificial neural network \citep{mnih2013playing}, where parameters are the weights of the network.
The algorithm calculates the estimator of the policy gradient and passes it to a gradient-based optimization algorithm like Stochastic Gradient Descent \citep{robbins1951stochastic, kiefer1952stochastic} or Adam \citep{kingma2014adam}.  
The equation for the basic policy \citep{Williams1992} gradient estimator is:
\begin{equation}
    \hat{g}(\theta)= \mathbb{E}_\theta [\sum_{i=1}^N\nabla_\theta \log \pi_\theta(a_i|s_i)\hat{R_i})], 
    \label{eq:expPolicyGrad}
\end{equation}
where $N$ is the total number of steps over all trajectories, and $\hat{R_i}$ is the return estimate from state $s_i$. The expectation $\mathbb{E}_\theta$ is approximated with a finite batch of sampled trajectories. 

We are using an upgraded policy gradient method called PPO \citep{schulman2017proximal, schulman2015trust} since original policy gradient methods are prone to catastrophically large policy updates. PPO limits the policy gradient updates to a "trust region" to prevent catastrophic fall in a performance that can be caused by large policy changes.

\subsection{AI Safety} 
An overview of AI safety methods can be found in \citet{pecka2014safe} and \citet{garcia2015comprehensive}. A large body of work in the area of Safe AI focuses on \emph{constrained RL} \citep{altman1999constrained,wen2018constrained}. Constrained RL depends on a-priori defined safety constraints which are states or actions that the agent should avoid. The constraints are often encoded within the environment reward functions. However, a challenge here is that when agents are transferred to a new task, learning again requires stochastic actions that could break safety. Additionally, there is the risk that the agent might forget its hazard avoidance skills during re-training on a new task. 

In other related work, \citet{lipton2016combating} introduced an approach in which a module is trained in a supervised way to predict the probability of catastrophic events. This module is integrated within the Q-learning objective. Another recent work that implements a similar modular idea of safety was introduced by \citet{srinivasan2020learning}, where a separate critic module is pre-trained on a random policy to approximate a Q-function that predicts a likelihood of hazard violation for an action. If the probability that an action will lead to a catastrophe exceeds a safety threshold, the probability to execute that action is set to 0. This approach is suitable for discreet action spaces. \citet{bharadhwaj2020conservative} provides an in-depth mathematical analysis of such approaches. Our work differs in that we in effect have two policies: the main policy and the instinct policy, both with their own action probability distributions and a modulation signal that can change the main policy's output. 
Furthermore, our approach is applicable to continuous action spaces.

In \citet{alshiekh2018safe} a system called "shield" monitors the agent's actions and overrides them if they would violate the pre-specified safety constraints. The safety constraints are specified through temporal logic. \citet{yuan2019modular} also employs goal specifications in temporal logic for safe RL. Other approaches to safe deep RL include estimating the safety of trajectories through Gaussian process estimation \citep{fan2019safety} or reducing catastrophic events through ensembles of neural models that capture uncertainty and classifiers trained to recognize dangerous actions \citep{kenton2019generalizing}. 

In this paper, we follow \citep{ray2019benchmarking} and
define an unsafe set of states as a subset of all states, $S_h \subset S$. The unsafe states represent situations or areas where we would like to minimize or completely prevent the agent from entering. For example, these states could represent walls that the agent should not collide with because of the resulting damage. 
Another example could be sidewalks and opposites lanes in the case of self-driving vehicles. While not always damaging, the car entering those areas exposes other traffic members to an increased risk. We would like to make sure that the car is avoiding those areas even when trying actions to learn a new task. We can define hazard violations as a binary variable $h(s_i) \in {0, 1}$, where $h(s_i)$ is $1$ if the current state is undesirable, and $0$ if it is not undesirable. We would like to minimize the expected violation return $V_{s_0} = \mathbb{E}[\sum_{s=s_0}^{s_T}h(s)]$ during the trajectory sampling, while still maintaining a reasonably good performance on the reward return $R_0$. The agent needs to know when it is in a hazards' neighborhood and to suppress the exploratory actions that can violate safety. Here, we assume that the subset $S_h$ stays the same across different tasks $\mathcal{T}_j$ and $\mathcal{T}_k$.

\section{Approach: Instinct Regulated Reinforcement Learning}
\label{section_approach}
Following the instinct network architecture introduced in \citet{grbic2020safe}, the agent's neural network is divided into two modules (Figure~\ref{fig_overview}a). The first module (policy network) has to learn the task, while the second module (instinct network) should learn to modify the main policy's actions when those would likely lead to safety violations. 

The instinct network is pre-trained to detect hazard zones and to engage instinctual actions to avoid them. Unlike related work  \citep{srinivasan2020learning}, this architecture is suitable for continuous action spaces. The final policy output is determined by: 
\begin{enumerate}
    \item Following the standard way of continuous action exploration in RL \citep{Williams1992}, the actions of the policy network $a^P$ are noisy; the policy network outputs a mean action $a_\mu$ that is given to a distribution (usually the normal distribution) from which the output action is sampled: $a^P_i \sim \mathcal{N}(a_\mu^n, \sigma^n)$, where $\sigma$ is part of the policy parameters $\theta^p$, and $n$ denotes $n^{th}$ action dimension.
    
    \item 
    The instinctual network is aware of the action $\vec{a}^P_i$ as well as the state observation $s_i$ at step $i$, creating the instinct state observation $s^I_i := \langle s_i, a^P_i \rangle$. This is in contrast to our previous MLIN approach \citep{grbic2020safe}, in which the instinct co-evolved to expect what kind of behavior the policy performs around hazards and therefore did not need $\vec{a}^P_i$ as input. 
     In our IR$^2$L approach, the instinct needs to work with a random policy on a task where hazards could be distributed differently  than during pre-training; the instinct needs to know what the policy wants to execute so it can modulate it accordingly.
    
    \item The instinct network outputs two instinctual actions: modulation value $m_i \in [0, 1]$ and the instinctual action $\vec{a_i}^I \in \mathcal{A}$.
    
    \item The modulation value $m_i$ is multiplied with the policy action vector $\vec{a}^P_i$ giving $\vec{a}^{P^*}_i$.
    
    \item  The instinct network action $\vec{a}^I_i$ is multiplied with $1 - m_i$ giving $\vec{a}^{I^*}$. If the policy action is getting suppressed by having $m_i$ close to $0$, the instinct action can pass through and vice versa. The final action $\vec{a}^F$ is the sum of $\vec{a}^{I^*}_i$ and $\vec{a}^{P^*}_i$.
    
\end{enumerate}

\section{Iterative Training Procedure}
\begin{figure*}[t]
\centering
\begin{subfigure}[t]{0.36\textwidth}
    \includegraphics[width=1.0\textwidth]{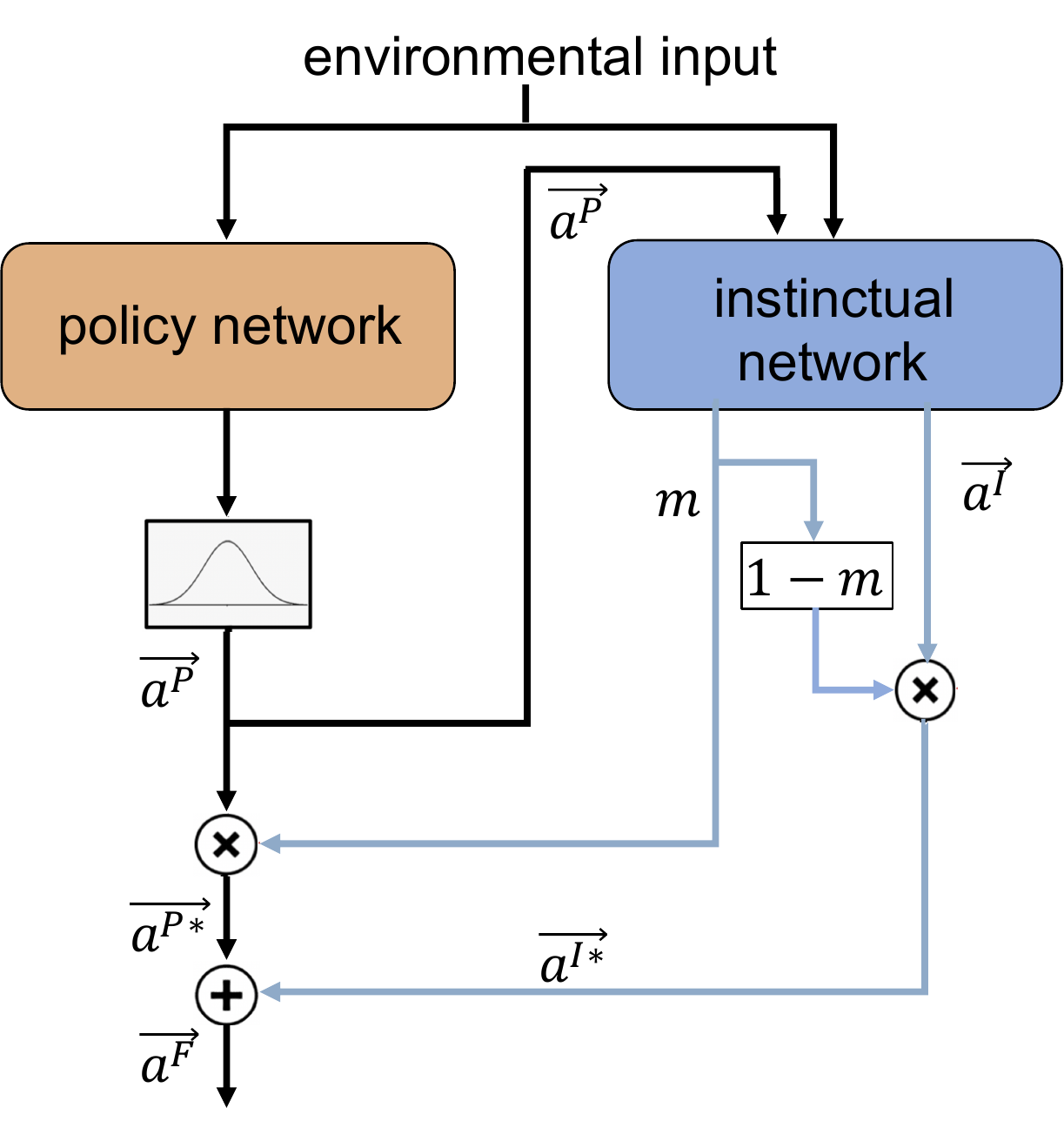}
    \caption{}
\end{subfigure}
\begin{subfigure}[t]{0.5\textwidth}
    \includegraphics[width=1.0\textwidth]{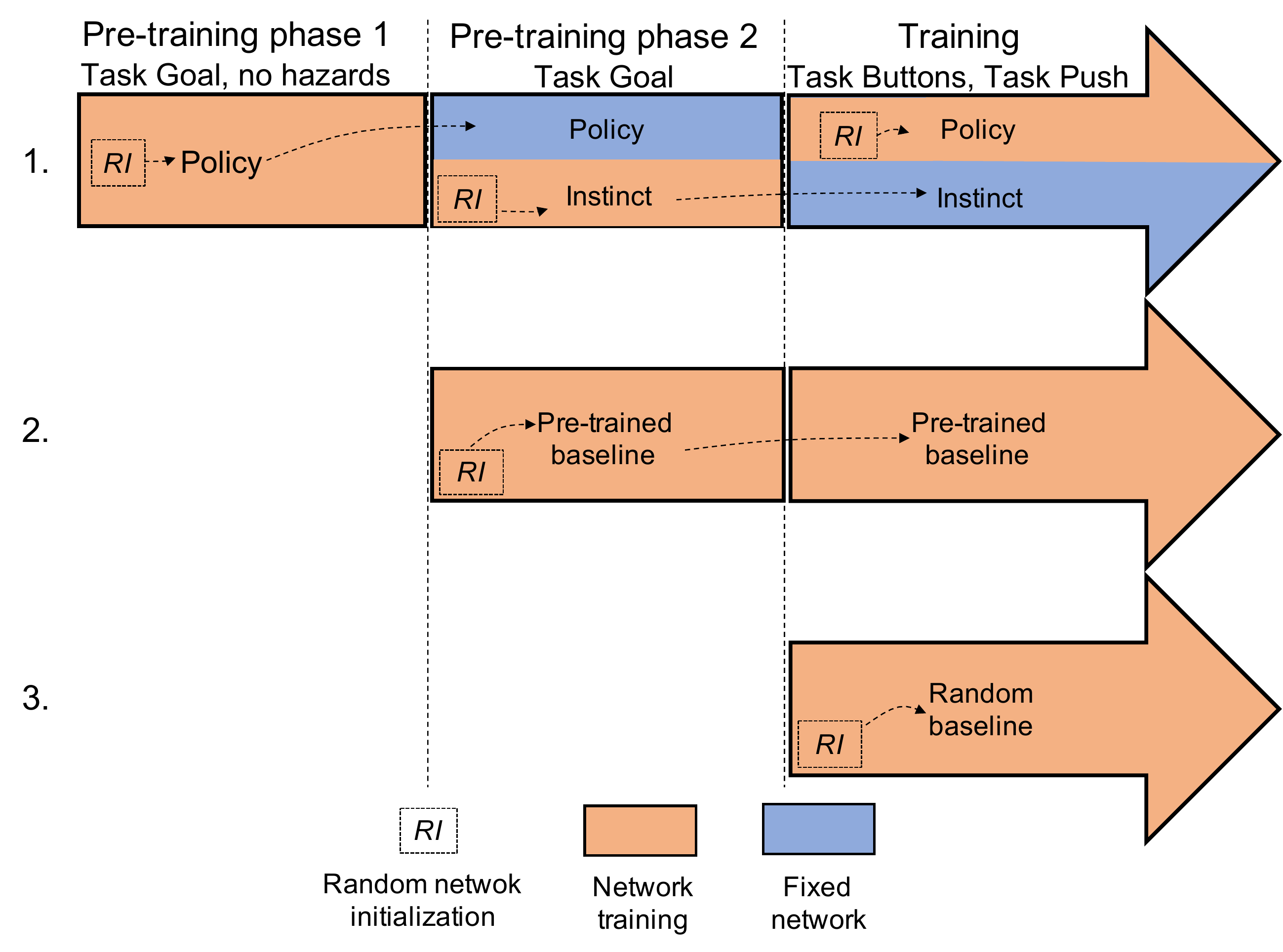}
    \caption{}
\end{subfigure}
\caption{Instinctual network architecture and instinct-regulated RL training procedure. \textbf{(a) The topology of the policy network with instinct module.} \normalfont Both networks receive the same input from the environment. The instinct network also takes the sampled policy action as additional input. The instinctual network outputs an instinctual action $\vec{a}^I$ and a suppression signal $m$. The suppression signal is a value between 0 and 1 that determines the magnitude of instinctual action that will be mixed in the policy action. The suppression signal $m$ is multiplied with policy action $\vec{a}^P$ and the opposite suppression signal $1-m$ is multiplied with the instinctual action $\vec{a}^I$. Two action values are finally added, resulting in the final action $\vec{a}^F$. \textbf{(b) Pre-training and training procedures for IR$^2$L and two baselines.} \normalfont \textbf{1.} IR$^2$L goes through two  pre-training phases (one for the helper policy and the other for the instinct) before it is transferred to the training tasks.  \textbf{2.} Pre-trained baseline is a single network adapted to the pre-training task and transferred to the training tasks. \textbf{3.}  Random baseline is a randomly initialized policy network exposed to the learning tasks.} 
\label{fig_overview}
\end{figure*}

Substituted the expansive meta-learning loop in MLIN \citep{grbic2020safe}, 
we present a more efficient sequential method where the instinct learns a transferable skill by being exposed to only one task.  We perform pre-training in two phases: (a) policy-only pre-training, and (b) instinct-only pre-training (Figure~\ref{fig_overview}b).

First, we train a policy on a task without any hazards and transfer the policy to the second phase consisting of the same type of task with added hazards. The policy is frozen, but an instinct is introduced that has to learn to prevent the policy from colliding with hazards. We consider the second phase to be a situation in which the agent can afford to commit many hazard violations during training (e.g.\ training in a simulator). The hypothesis is that after this pre-training phase, safe learning from a random policy in safety-critical domains should be possible by combining this policy with the hazard-avoiding skills of the instinct.   
We assume that the hazard observations are invariant between all tasks.
In more detail, the pre-training phases include:  


\textbf{Phase 1: Policy-only pre-training without hazards.} 
We train a policy network without an instinct component to solve a task without any hazards in the environment. 
 The purpose of this phase is to have a  policy that can perform the task well but not safely. When transferred to the instinct pre-training phase, it should help the instinct to collect relevant hazard experiences.

\textbf{Phase 2: Instinct pre-training.}  
%
The second pre-training phase introduces hazards to the task used in the first pre-training phase. The policy from the first pre-training phase is expected to do frequent hazard collisions and thus provide abundant experiences for an instinct to learn to avoid. During this phase, the policy is fixed to the policy found in the first pre-training step 
while the instinct is randomly initialized. Unlike during later training of the policy during transfer to test tasks ("Training" column in Figure \ref{fig_overview}b) in which a  random policy executes stochastic actions and the instinct deterministic ones, here the instinct executes stochastic actions and the policy executes deterministic ones.
The goal of the second phase is to train an instinct that stops an agent from colliding with hazards while still giving the policy enough flexibility to reach its goal. To achieve this task, we designed the following instinct reward function:
\begin{equation}
    r_{i}(s_i, \vec{m}_i, r_t, h)= (1 - h(s_i) H)m_i r_t(s_i) D,
    \label{eq:instinctReward}
\end{equation}
where $s_i$ is the current state observation, $m_i$ is the instinct suppression value, $r_t$ is the reward given for getting closer to the goal. $h(s_i) = 1$ if the agent collided with a hazard in the $i$-th step, otherwise $h(s_i) = 0$ . The first term $1-h(s_i)H$ models the safety, and is $1$ if the agent is safe. If the agent collided with a hazard, the term becomes smaller for $H > 0$. If the hyperparameter $H > 1$ the safety term will become negative and the instinct reward will be negative overall. The hyperparameter $H$ controls how harsh  the instinct will be punished for not preventing collisions with hazards and, as a result, how conservative it will be in the end.

As a reminder, $0 < m_i < 1$, where $m_i=1$ means that only the policy is controlling the agent, and $m_i=0$ means that the instinct took complete control. The smaller $m_i$ is, the smaller is the instinct reward, discouraging the instinct from activating when not necessary. Since the instinct works in combination with a policy that is efficient in reaching the goal, the goal-reaching reward $r_t$ is there to maximize the reward when the instinct is not active. We would like to discourage the instinct from doing anything if the agent operates within safe parts of the environment, activating only if the agent is in danger of colliding with a hazard. Hyperparameter $D$ is there to amplify the contribution of $r_t$. We performed a hyperparameter search on $H$ and $D$ and found that $H = 100$ and $D = 15$ work best for our experiments. The instinct was trained for 300 epochs and in each epoch, 215 trajectories were sampled.



\section{Task environment}
We test our approach on the OpenAI Safety Gym environment developed exactly for studying reinforcement learning for safe exploration \citep{ray2019benchmarking}. The framework provides a variety of tasks, agents, and obstacles that can be easily rearranged and modified and is built upon the MuJoCo physics engine for robotics simulations.

\begin{figure*}[t]
\centering
\begin{subfigure}[t]{0.3\textwidth}
    \includegraphics[width=1.0\textwidth]{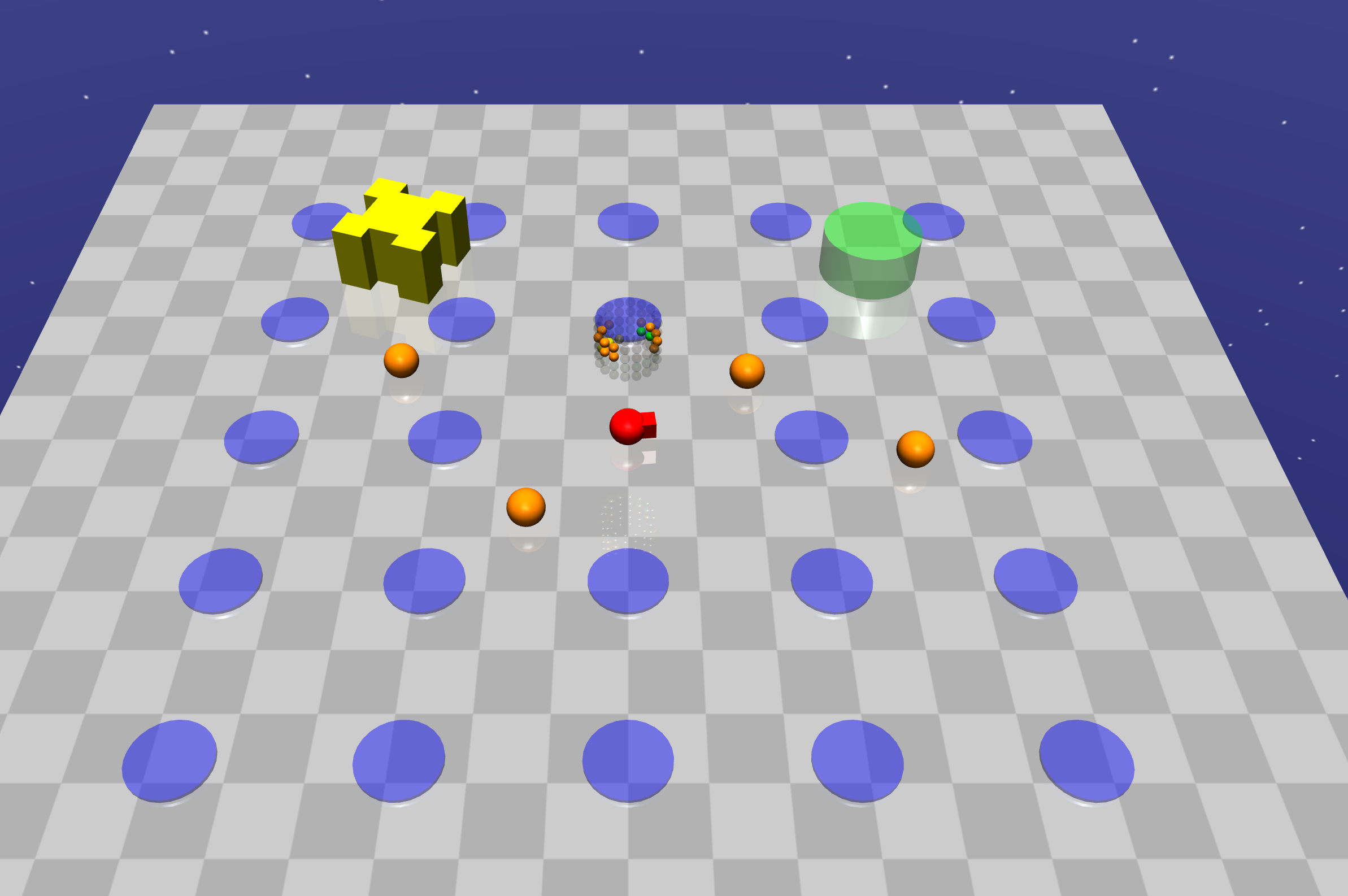}
    \caption{Task Goal}
\end{subfigure}
\begin{subfigure}[t]{0.3\textwidth}
    \includegraphics[width=1.0\textwidth]{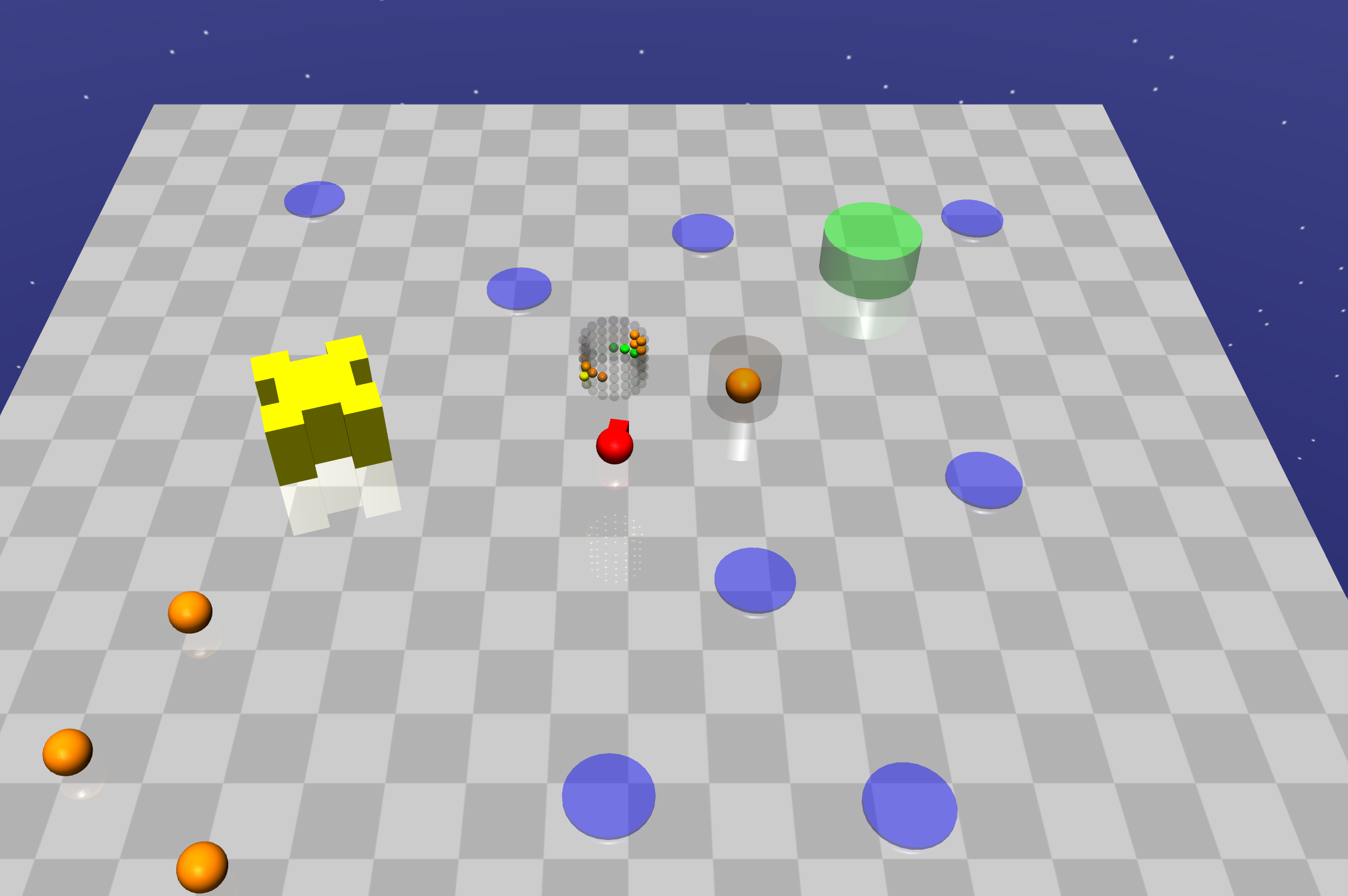}
    \caption{Task Buttons}
\end{subfigure}
\begin{subfigure}[t]{0.3\textwidth}
    \includegraphics[width=1.0\textwidth]{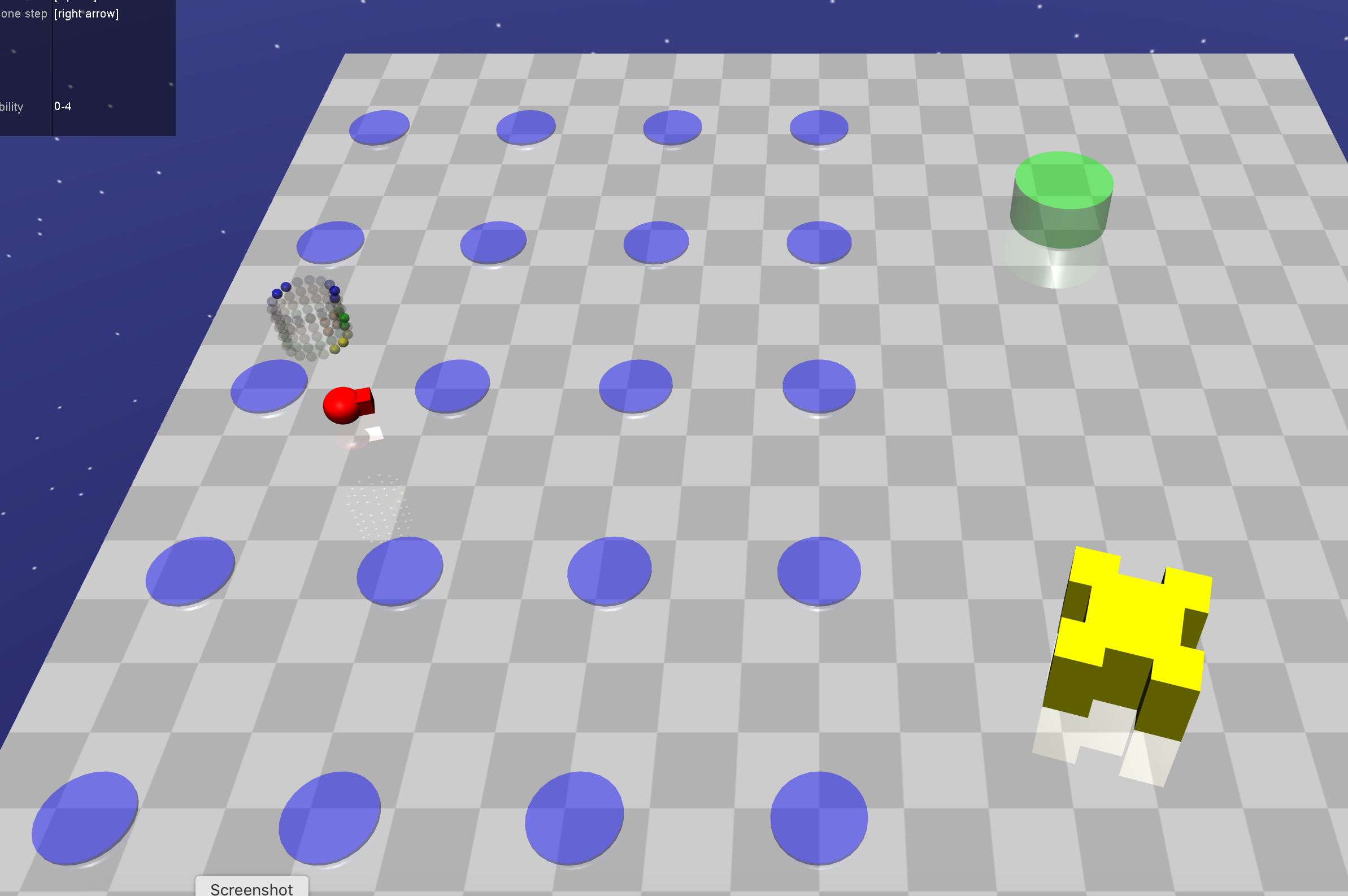}
    \caption{Task Push}
\end{subfigure}
\caption{\textbf{(a) Task Goal:} The agent has to reach the green cylinder while avoiding hazards (blue circles). As soon as the agent reaches the green cylinder, the cylinder reappears on a different, randomly chosen, spot. \textbf{(b) Task Buttons:} There are 8 randomly distributed hazards. The agent has to reach one of the 4 randomly distributed buttons. When the agent reaches the correct button, one of the other buttons becomes the correct one. The correct button is highlighted with a gray halo. \textbf{(c) Task Push:} There are 20 hazards distributed in a 5$\times$4 grid. The agent spawns in the area left of the hazard grid and has to clear the hazards grid to push the box into the goal.} 
\label{fig_tasks}
\end{figure*}

The environment allows three different task types: "goal", "push", and "buttons" (Figure~\ref{fig_tasks}).
In the \textbf{"goal"} task type, the agent (red body) has to reach a green cylinder randomly spawned in the environment. In the \textbf{"buttons"} task type there are several buttons (orange spheres) where the agent needs to learn to press the correct button. The most complex \textbf{"push"} task type requires the agent to push a yellow box in the green cylinder. The agent is challenged by a variety of obstacles that generate cost if the agent collides with them. For the purpose of this paper, we focused only on static "hazard zones" (blue circles) that the agent can cross but that generate a cost for every step the agent spends within one of them.

The agent is equipped with a set of lidars for detecting environment elements. There is a separate set of lidars for each element (colored crowns above the agent in Figure \ref{fig_tasks}). Each lidar set has 16 lidars distributed around the agent at equal angles. They are represented as a vector with each lidar observation in the range [0, 1], 0 when the target is not visible and 1 when the observed object is adjacent to the agent. Furthermore, the agent has access to its orientation relative to the north and the central point of the map. 

The agent is a floating dot that only moves on a two-dimensional surface. The available actions are represented by a two-dimensional vector where the first dimension represents backward/forward movement, while the second dimension represents left/right turning movement. 

In the original Safety Gym setup, 
the agent sees the goal object and the correct button with the same lidar set. Thus, a policy trained to follow the green cylinder in the "goal" task type, would  perfectly transfer to the "buttons" task type. To make it more challenging for the agent, we modified the original Safety Gym code to separate the lidar set into two different sets, one for "buttons" and one for "goal" task types. We also modified the framework to allow different lidar ranges for hazard detection and for task elements (box, cylinder, button). If the original range was too short, the agent could not see goals too far away. If the lidar range was too long, the agent would get over-saturated with hazards lidar inputs in case there were a lot of hazards present in the environment. We modified the original source code to allow for short lidar ranges when detecting hazards, and long-range when detecting other elements. The episode length of all implemented tasks is equal to 1000 steps ($s_T=s_{1000}$).

\textbf{Task Goal.} 
Here the agent has to reach the green cylinder randomly placed on the map (Figure \ref{fig_tasks}a). The agent is spawned in the center of the map at the start of each episode. If the agent reaches the goal before the episode finishes, the goal is placed at a different location on the map. There are other elements on the map (yellow box and orange buttons) but they are irrelevant to this task, so the agent needs to learn to ignore them. The hazards are distributed in a static 5$\times$5 grid, with the central hazard missing, equaling to 24 hazards in total. Since this task is used only in the instinct pre-training phase, we wanted to maximize the opportunities where the agent can collide with the hazard while still being able to learn to reach the goal.

\begin{figure}
\centering
\begin{subfigure}[t]{0.45\textwidth}
    \includegraphics[width=1.0\textwidth]{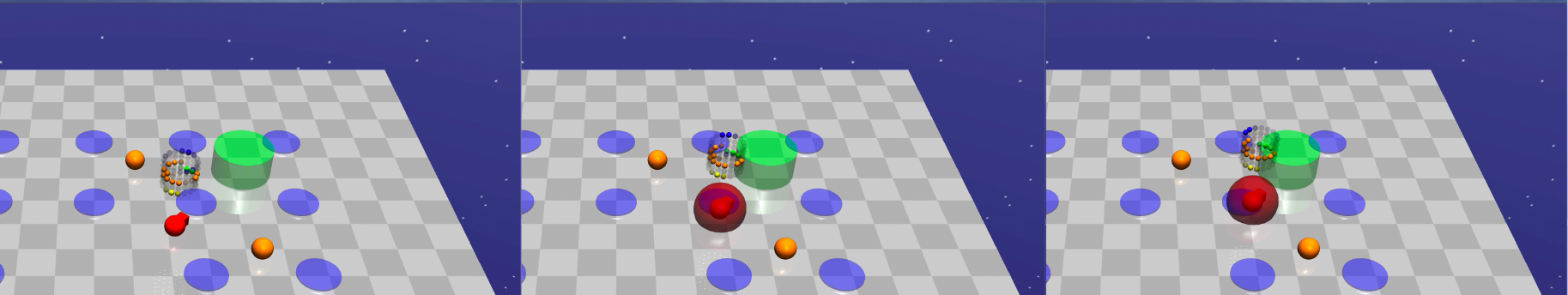}
    \caption{}
\end{subfigure}
\begin{subfigure}[t]{0.45\textwidth}
    \includegraphics[width=1.0\textwidth]{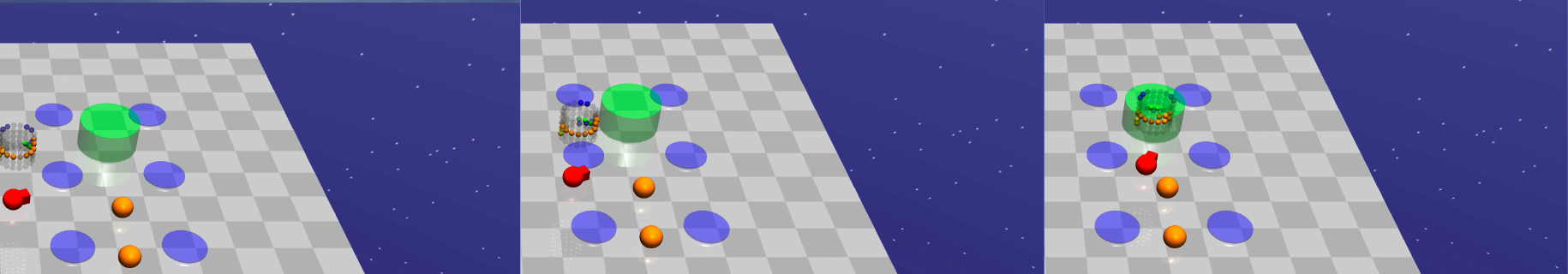}
    \caption{}
\end{subfigure}
\caption{\textbf{(a)} A sequence of frames showing policy from pre-train phase 1 going over a hazard in Task Goal. \textbf{(b)} The same policy working in conjunction with an instinct trained to avoid hazards in Task Goal. The instinct can learn to regulate unsafe policy actions.} 
\label{fig_Inst_NoInst}
\end{figure}

The reward is defined as the negative difference between the agent-goal distance in this step and in the previous step. If the agent came closer to the goal, the reward is positive. Otherwise, the reward is negative. For each step that the agent spends in a hazard zone, the environment hazard function $h(s)$ returns $1$, otherwise, it returns $0$. The behavior of the policy trained without hazards and the same policy after we added the pre-trained instinct can be seen in Figure \ref{fig_Inst_NoInst}.

\textbf{Task Buttons.} 
The agent is spawned in the center and has to press the correct button out of four randomly positioned buttons (Figure~\ref{fig_tasks}b). All four buttons are detectable through one lidar set, and the correct button is visible through a separate lidar set. The buttons are fixed throughout the episode and the next correct button is randomly chosen as soon as the agent presses the current correct one.

There are eight randomly positioned hazards at the beginning of an episode. This is 
the first task used to test the hazard avoidance capabilities of the instinct network. We would want the agent to learn to press the correct buttons while avoiding the hazards during the training phase. The reward is the measure of how much closer the agent is to the correct button since the last step. The $h(s)$ function communicates whether the agent is stepping over a hazard.

\textbf{Task Push.} 
The goal of the agent is to push the yellow box into the goal (green cylinder) (Figure~\ref{fig_tasks}c). The agent gets a reward for getting close to the yellow box and another reward for closing the distance between the box and the goal. The two rewards are summed, resulting in the final step reward. The hazards are spawned in a 5$\times$4 grid on the left-hand side, while the box and the goal are located on the right-hand side of the map. The agent is always spawned just at the left of the hazards grid.

The hazards are fixed in a grid layout while the goal and the box are randomly spawned in their area. The reasoning for this layout was that in an environment where hazards, box, goal, and the agent are uniformly distributed across the map, it is extremely difficult for the agent to push the box around the hazards. 
For that reason, we made it easier for the agent to solve the task by decoupling the obstacles from the task. The agent needs to clear the hazards grid to solve the task. The environment hazard function $h(s)$ communicates hazard violations.

\section{Training details}

\subsubsection{Network implementation details.}
For the \emph{policy network} and the \emph{instinct network}, we use an advantage actor-critic system \citep{konda2000actor}, where actor and critic are two separate, fully connected neural networks with three hidden layers of 512 neurons each and Tanh activation functions. The policy gradient in the advantage actor-critic system can be described as:  
\begin{equation}
    \hat{g}(\theta) = E_\theta [\nabla_\theta \log f_\theta(s,a)A_\theta(s)], 
    \label{eq:policyGrad}
\end{equation}
where $A_\theta(\cdot, \cdot)$ is the advantage calculated from the critic and $f_\theta(\cdot, \cdot)$ is the output of the actor-network.
The critic-network is updated to minimize the temporal difference between predicted expected return $A_\theta(s_t)$ at state $s_t$, and the reward $R(s_t)$ updated return estimate: $A_\theta(s_t) - (R(s_t) + \gamma A_\theta(s_{t+1}))$, where $\gamma$ is the reward discount hyperparameter \citep{peters2005natural, wu2017scalable}.

 The actor outputs a mean action  for a Gaussian distribution $\mathcal{N}(\vec{a}_\mu, \vec{\sigma})$,  
 from which an action is sampled \citep{Williams1992}. The critic outputs the predicted value (predicted future cumulative reward). The final layer of the plastic policy’s actor-network has two outputs (Tanh) scaled to $[-0.1, 0.1]$, reflecting the Safety Gym Point agent's action space. The instinct actor-network outputs three outputs; two are the instinct actions reflecting the agent action space and the last one is the instinct modulation signal $m$. The modulation signal is then scaled to fit the $[0, 1]$ range. We clip the modulation signal to $[0, 1]$ range in case sampling from a Gaussian distribution $\mathcal{N}(\vec{a}^I_\mu, \vec{\sigma}^I)$ takes it outside the range. 

\subsubsection{RL training details.}

The weights of all networks implemented for this paper are initialized with Kaiming uniform initialization \citep{He_2015}. Gaussian action noise parameter $\sigma$ is initialized to 0.6 and the learning rate is set to 0.001. In each training epoch, the sample buffer collects 216,000 state-action-reward samples which are equal to 216 trajectories/episodes.
An existing PPO implementation from \cite{pytorchrl} served as a starter code for our implementation of the method. There is a set of PPO hyperparameters needed for training: $\gamma$ discount factor (0.99), PPO clip parameter (0.2), PPO epoch number (4), value loss coefficient (0.5), and entropy term coefficient (0.01).
This configuration was used for both pre-training phases and experiment task adaptation; The implementation of our method can be found at \url{github.com/djole/IR2L}.

\section{Task Transfer Results} 
\label{sec_task_transfer_res}

\begin{figure}[t]
\begin{center}
\includegraphics[width=0.45\textwidth, angle=0]{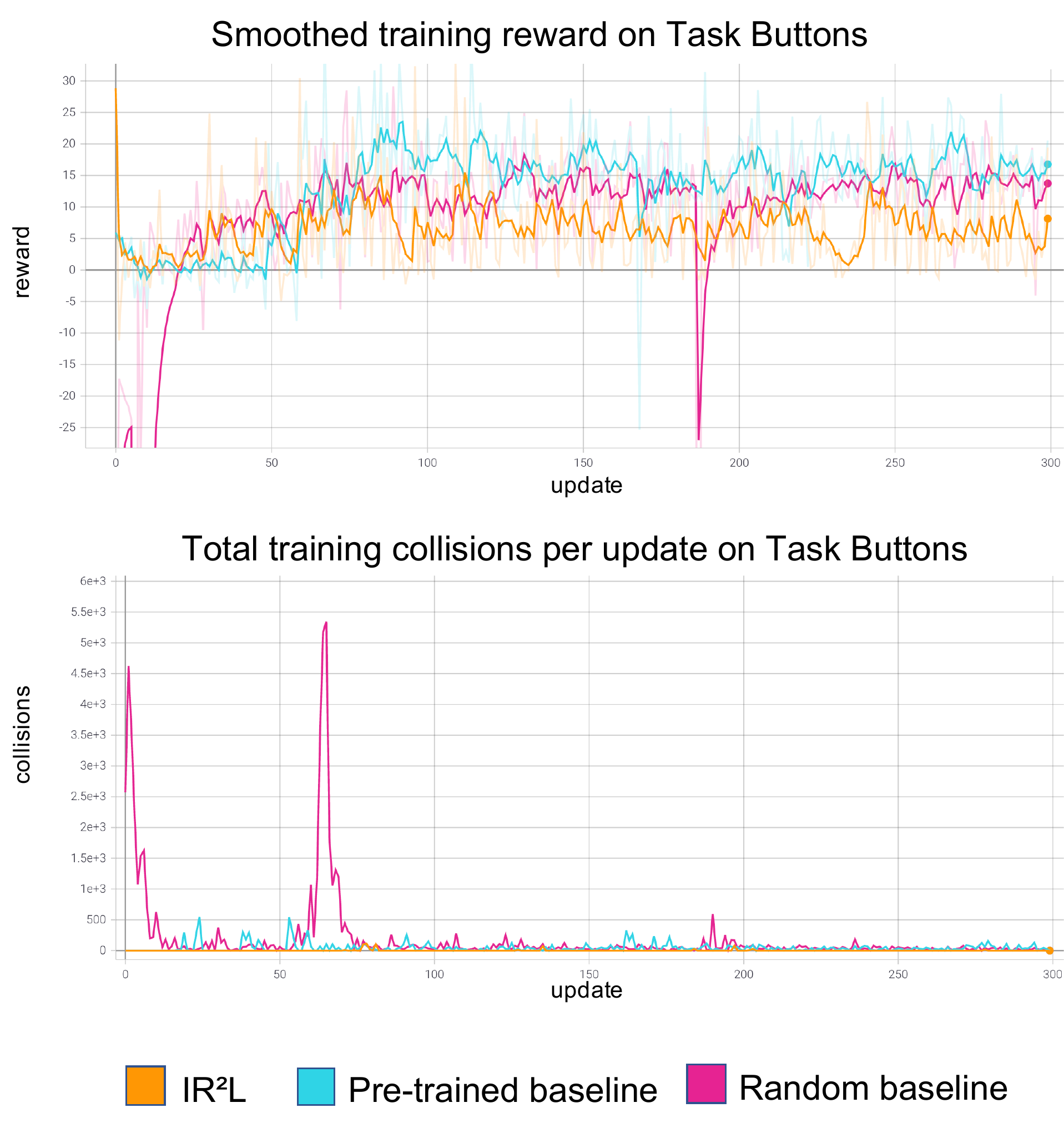}
\vskip 0.25cm
\caption{Reward and collisions during training in Task Buttons. After each episode, we evaluate the agent on a single episode and plot the cumulative reward. The collisions are added over all exploratory episodes per update. All methods show comparable task rewards with IR$^2$L consistently avoiding hazards throughout  training.
}
\label{fig_buttons_training_reward}
\end{center}
\end{figure}
We compare the final task reward and the number of collisions during training between IR$^2$L that uses a pre-trained instinct module  and two baselines that do not use an instinct module. The first
is a \textbf{random baseline}, i.e.\ a randomly initialized policy that has to learn Task Buttons and Task Push from scratch. The second one represents a  baseline that is  \textbf{pre-training} on Task Goal and then transferred to the other two tasks. The idea is that pre-training a policy, even without an instinct network, should reduce safety violations when learning another task afterward. 
%
The task reward during experiments is
$r_t^*(s) = r_t(s) - h(s)H_t$,
where $r_t(s)$ is the task-type specific reward, and $h(s)$ is a binary function indicating hazard violations. Hyperparameter $H_t$ is a task-specific punishment for colliding with a hazard. The hyperparameter $H_t$ was chosen to optimize baseline's learning to avoid hazards while still being able to solve the tasks. We found that $H_t = 1$ and $H_t = 10$ works best for baseline training on Task Buttons and Task Push respectively. We used the same task reward for baselines and IR$^2$L. Videos of the resulting behaviors can be found: \url{https://youtu.be/lqRvHimqvAc}.

\textbf{Transfer to Task Buttons.} 
We trained the baselines and IR$^2$L for 300 training epochs with each epoch having 215 sampled trajectories in the training buffer (Figure \ref{fig_buttons_training_reward}). This setup leads to a significant reduction in training collisions. In the case of the baseline,  the noise in the training curves is due to reward punishments caused by colliding with hazards, while in the case of the 
instinct they can be the results of an agent unable to  move,  with hazards  blocking the path to the correct button. The rewards are measured on a single episode with a deterministic policy after a weight update, while collisions are measured during the sample collection before the update.

\begin{figure}[t]
\centering
\begin{subfigure}[t]{0.23\textwidth}
    \includegraphics[width=1.0\textwidth]{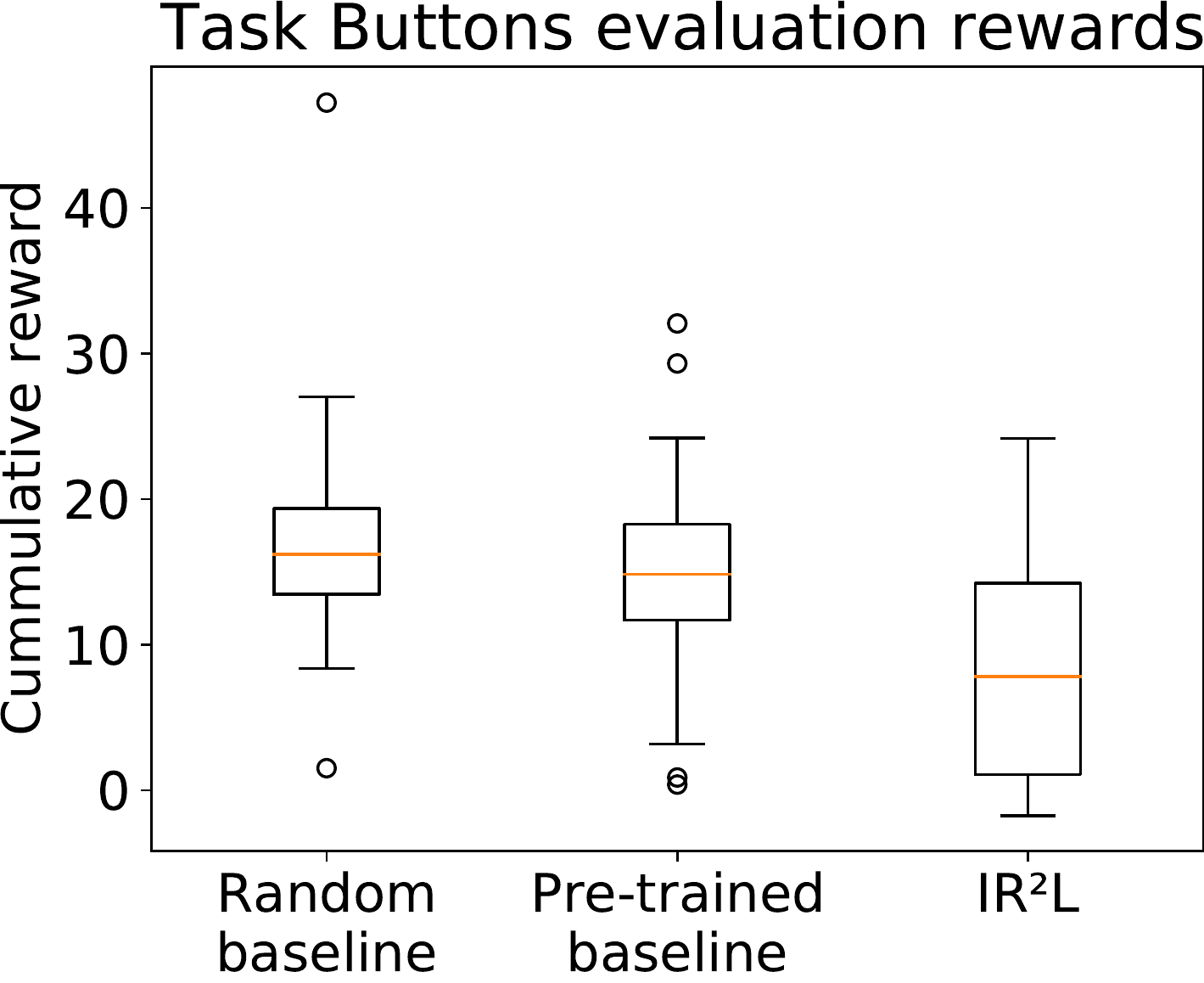}
    \caption{}
\end{subfigure}
\begin{subfigure}[t]{0.23\textwidth}
    \includegraphics[width=1.0\textwidth]{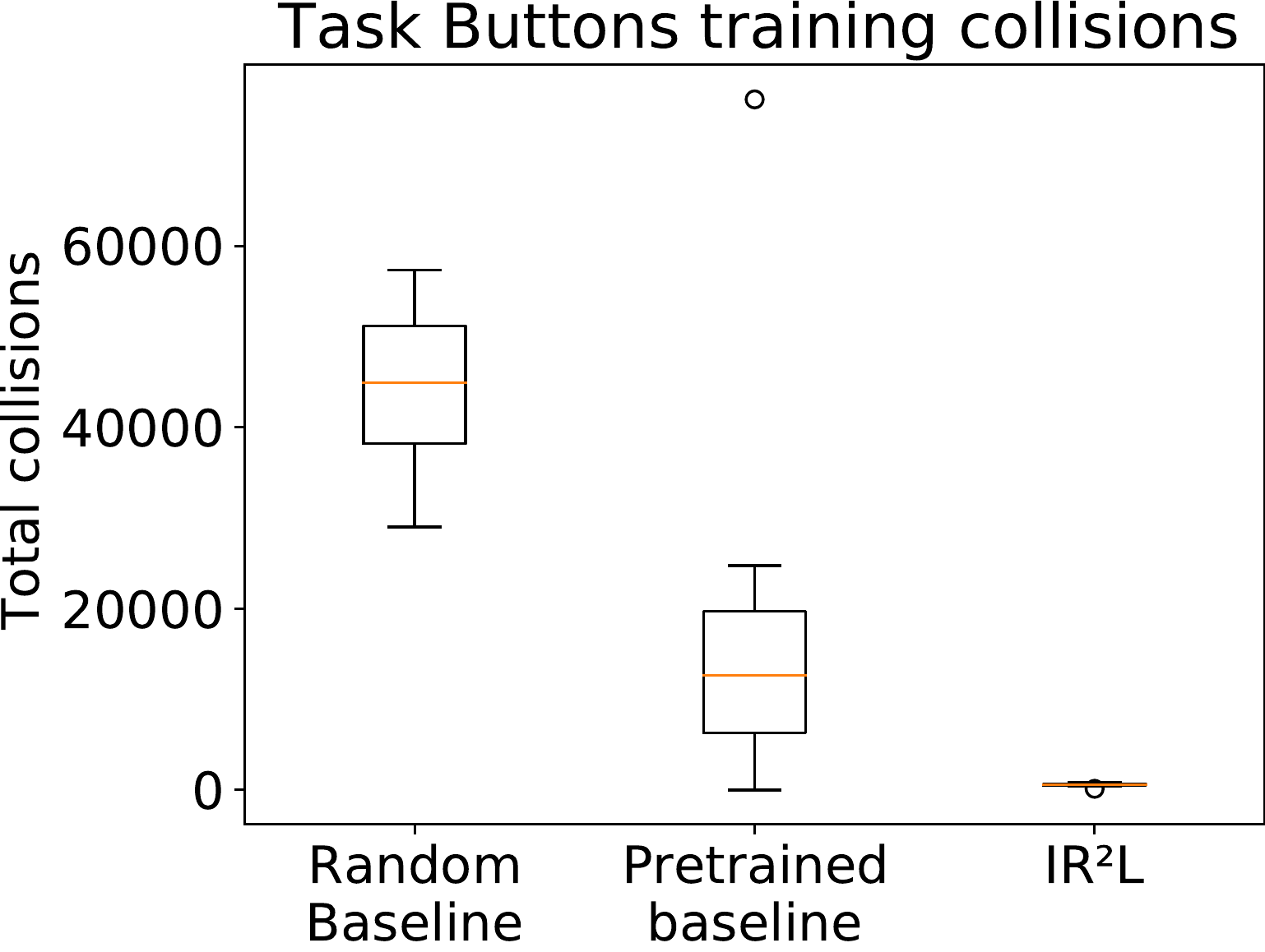}
    \caption{}
\end{subfigure}
\caption{\textbf{(a) Reward: }Box-plot showing the cumulative reward on 50 episodes for the final baselines and $IR^2L$ on Task Buttons. \textbf{(b) Training collisions: } Box-plot showing the cumulative collisions on 10 training runs. Median collision number for IR$^2$L is 608. IR$^2$L shows large reductions in training collisions while still being able to perform the task.} 
\label{fig_button_results}
\end{figure}
The reward calculations during training  (Figure~\ref{fig_buttons_training_reward}) also include the hazard violations. To better distinguish the performance on the task and the ability to avoid hazards, we evaluate  the models on 50 episodes purely on the task rewards (hazards not accounted for).  
%
The baseline agents move faster than the IR$^2$L agent, which is moving more carefully and thus receives a slightly reduced task return  (Figure~\ref{fig_button_results}a).

In the original Safety Gym paper \citep{ray2019benchmarking}, the benchmarks for the vanilla "buttons" type task show cumulative rewards to be ~25 on average with 200 collisions per episode on average for unconstrained methods (the task reward here does not include hazards punishments). Constrained methods show the cumulative reward to be between 0 and 5 on average, while collisions to be 25 per episode on average during training.

\begin{figure}[t]
\begin{center}
\includegraphics[width=0.45\textwidth, angle=0]{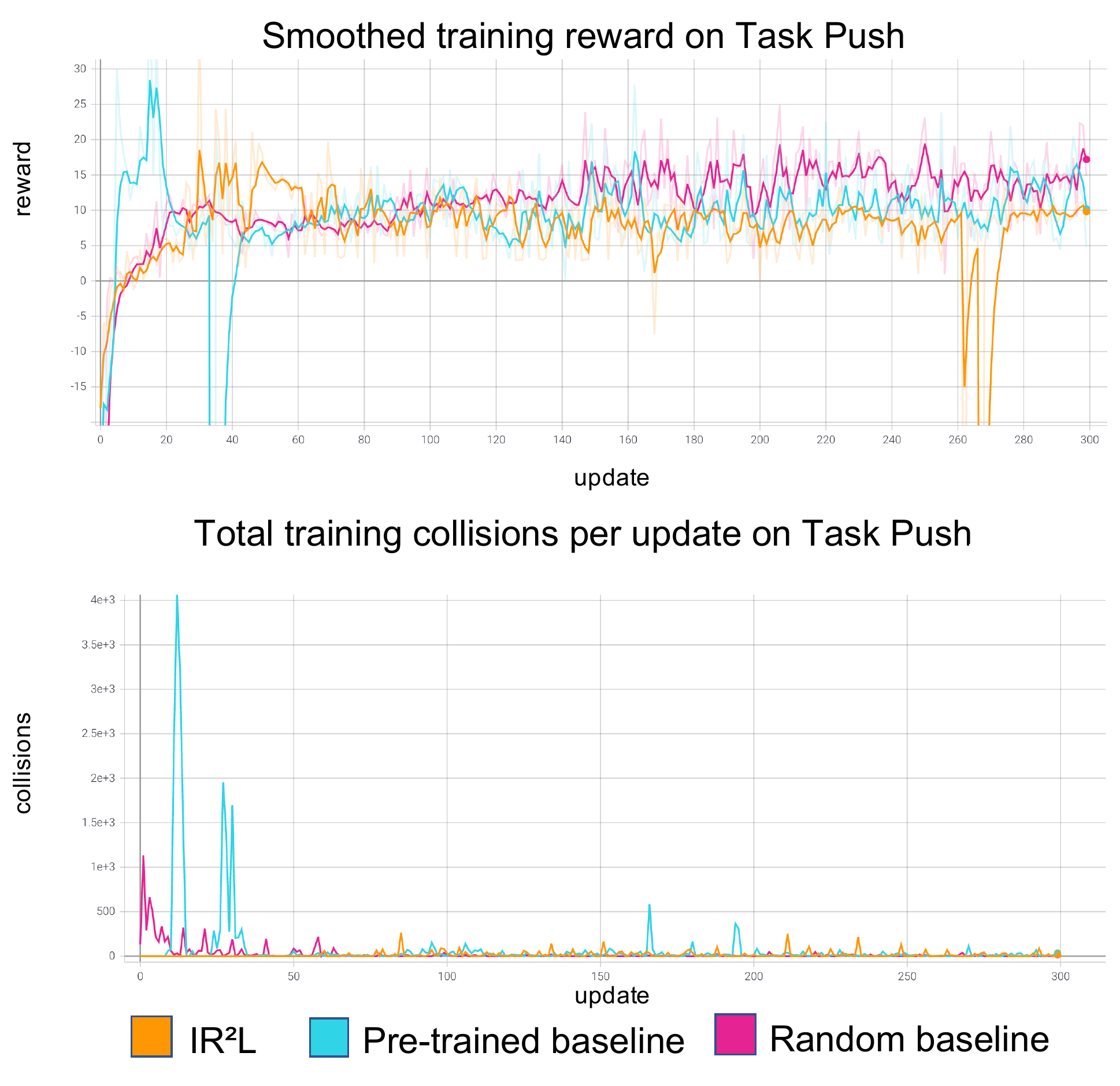}
\vskip 0.25cm
\caption{Task Button rewards and collisions during training. 
The approaches display a comparable task reward with IR$^2$L consistently avoiding hazards throughout  training, while both baselines show large collision spikes at the start of training.
}
\label{fig_push_training_reward}
\end{center}
\end{figure}

We repeated the Task Buttons experiment 10 times and plotted the cumulative hazard collisions during the learning period for the baseline policies and IR$^2$L (Figure~\ref{fig_button_results}b).  Due to the stochastic nature of the task and imperfect instinct training during the pre-training phase, IR$^2$L performs some hazard violations but shows substantial improvements over the two baselines. As expected, the pre-trained baseline shows a strong transfer of hazard-avoiding skills to the new task compared to the random baseline. 



\textbf{Transfer to Task Push.} 
The number of training epochs and samples is the same as for Task Buttons (Figure~\ref{fig_push_training_reward}).  Training with instinct protection allows the policy to almost completely avoid hazards. Not surprisingly, the random baseline displays a large spike of hazard collisions at the start of training until it start learning to avoid them. The pre-trained baseline also shows large spikes in collisions during training since there is nothing to protect a stochastic policy from stepping over hazards in the dense grid of hazards present in Task Push. The task reward over training steps shows a similar performance between all methods. Large dips in task reward are sometimes observed during training, which is due to the agent occasionally stepping over hazards and receiving a punishment of $H_t = 10$. 
The dips are not visible in the "collisions" plot since the rewards are measured on a single episode with a deterministic policy after a weight update, while collisions are measured during the exploration phase before the update. The Safety Gym  benchmarks  \citep{ray2019benchmarking} for the vanilla "push" type task show cumulative rewards to be ~7 on average with 40 collisions per episode on average. Constrained methods show the cumulative reward to be ~2 on average, while collisions to be ~25 per episode on average during training. The "push" task implementations in Safety Gym and here are significantly different so the numbers are not perfectly comparable.

\begin{figure}[t]
\centering
\begin{subfigure}[t]{0.23\textwidth}
    \includegraphics[width=1.0\textwidth]{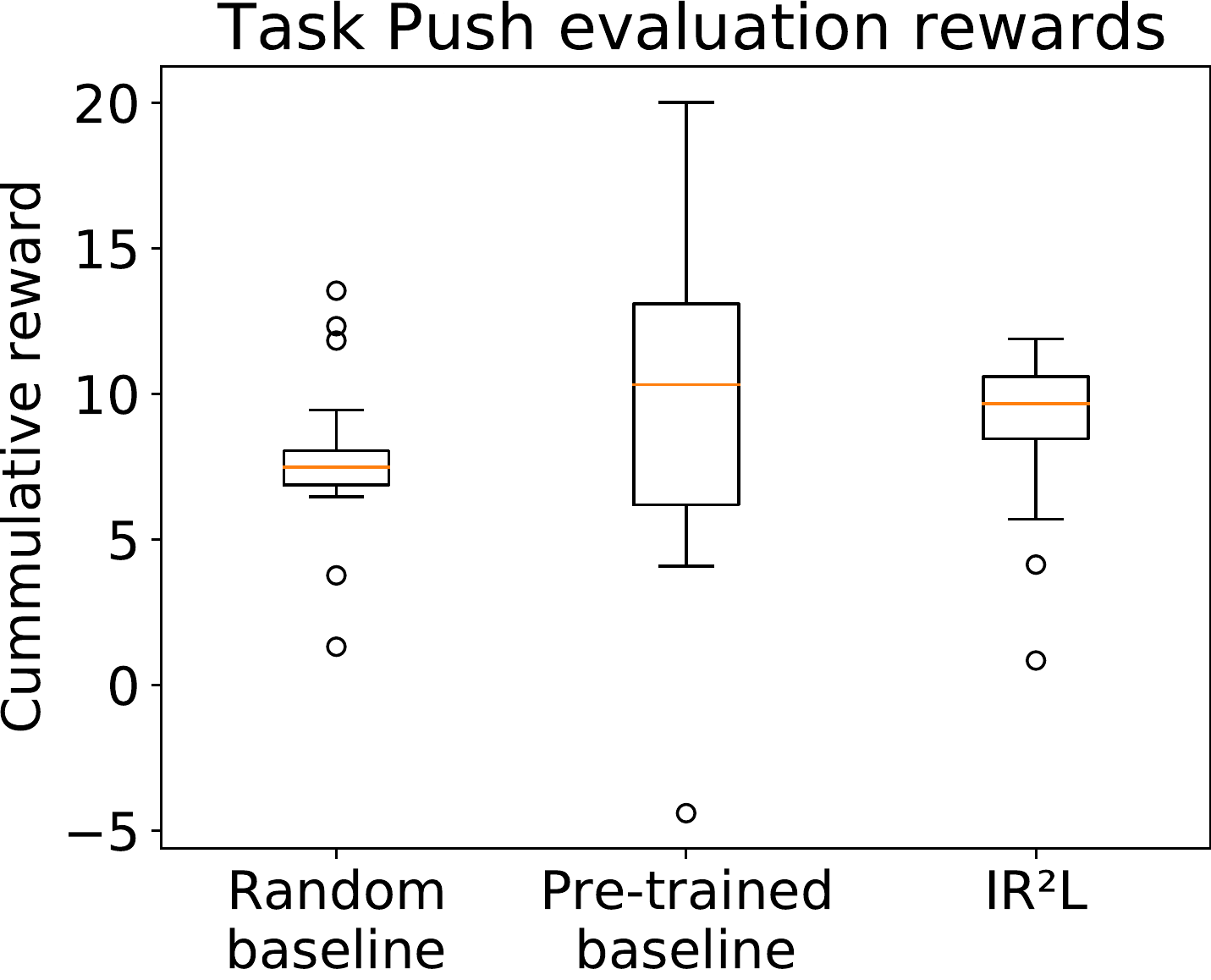}
    \caption{}
\end{subfigure}
\begin{subfigure}[t]{0.23\textwidth}
    \includegraphics[width=1.0\textwidth]{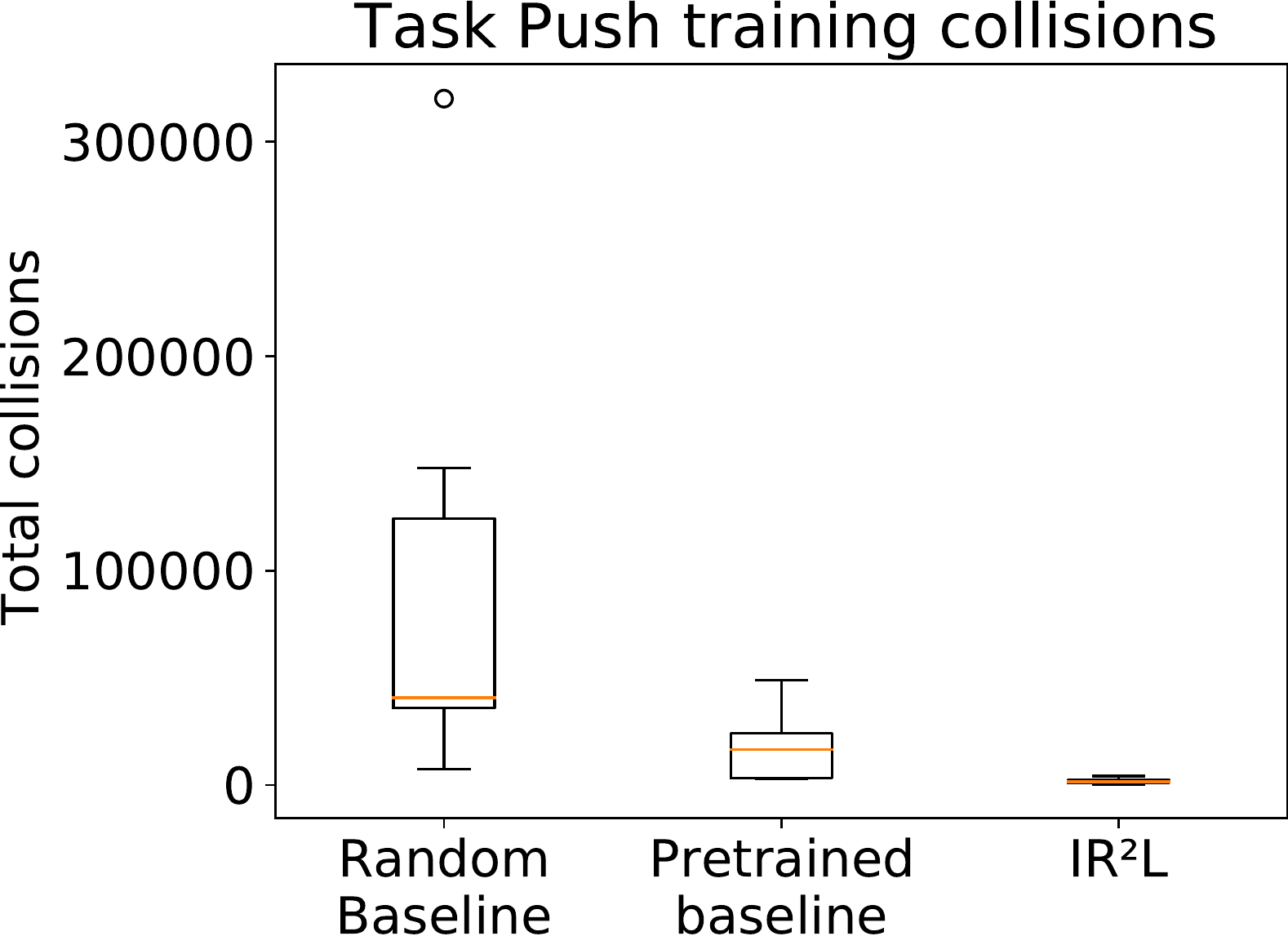}
    \caption{}
\end{subfigure}
\caption{\textbf{(a) Reward: }Box-plot showing the cumulative reward on 50 episodes for the final baseline and IR$^2$L on Task Push. \textbf{(b) Training collisions: } Box-plot showing cumulative collisions during 10 training runs. $IR^2L$ has a significantly smaller number of collisions during training with similar task learning capabilities than the baselines.} 
\label{fig_push_results}
\end{figure}

The result of 50 evaluation episodes of the final policy is shown in Figure~\ref{fig_push_results}a. The IR$^2$L approach shows a similar final median performance on the task reward
as the baselines. The pre-trained baseline is showing better performance in some episodes compared to the random baseline and IR$^2$L, 
likely due to a large skill transfer between Task Goal and Task Push. We also repeated the Task Push experiment 10 times and plotted the cumulative hazard collisions during learning for the baseline policies and IR$^2$L (Figure~\ref{fig_push_results}b). The pre-trained baseline has a strong hazard-avoiding and goal-reaching skill transfer. Still, the IR$^2$L method has substantially 
better hazard-avoiding skills while maintaining similar reward returns on average. 

\begin{figure}[ht]
\begin{center}
\includegraphics[width=3.0in, angle=0]{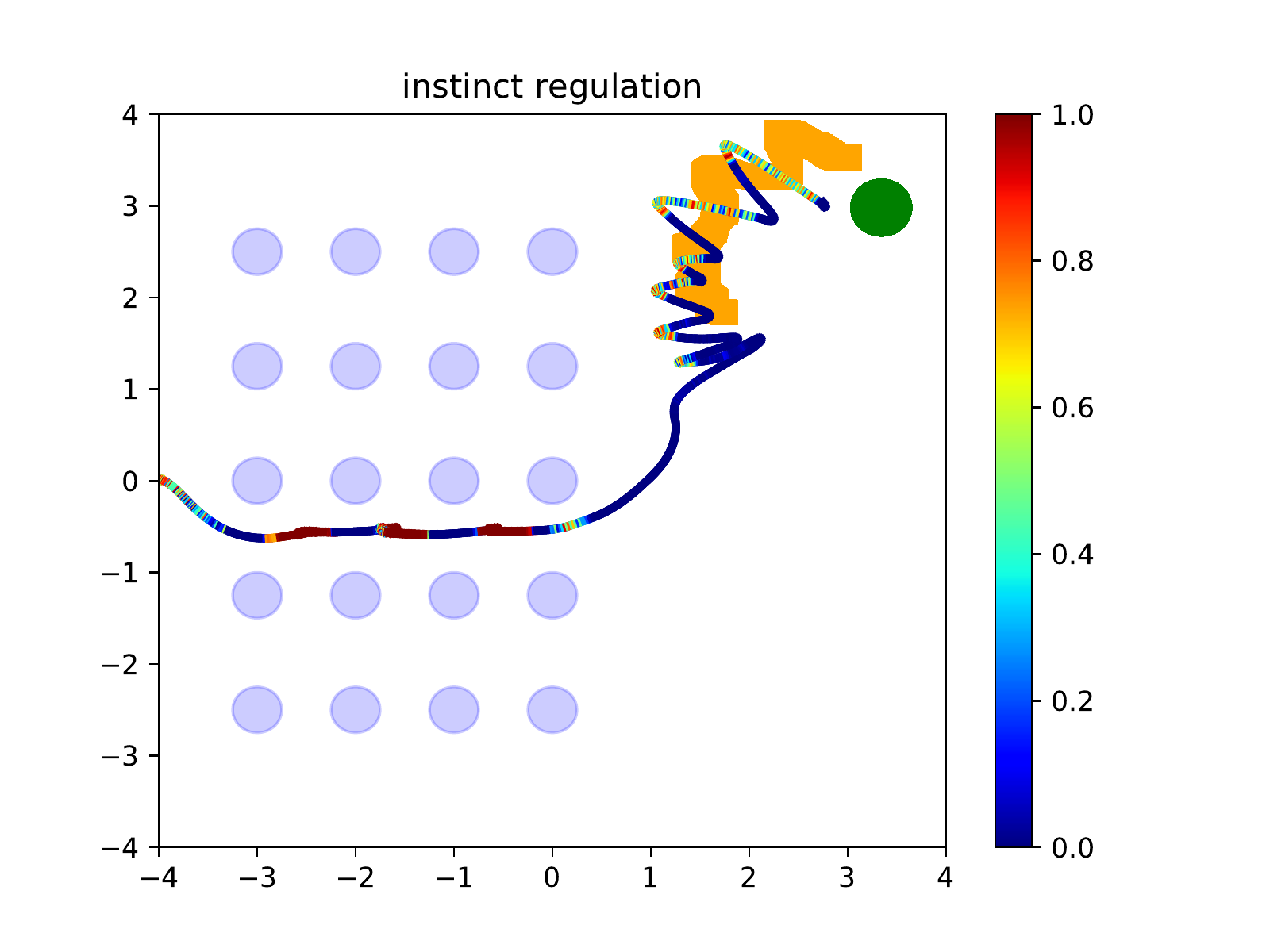}
\caption{Agent trajectory on the Task Push. The more red a line segment, the more the instinct is active.  
Thick orange lines indicate the box trajectory, while the  green circle indicates the goal location.
}
\label{fig_push_trajectory}
\end{center}
\end{figure}

We observe that the random baseline policy is very quick to clear the hazards grid and reach the box and the goal on the other side but has a much higher risk of hitting the hazards. The instinct-protected policy is, on the other hand, more careful around the hazards grid. Figure~\ref{fig_push_trajectory} shows a trajectory of the agent with high instinct activation within the hazards grid, keeping the agent safe from collisions. Even with the relaxation of the task with separate hazard and box/goal zones, the task is challenging to learn for the baseline and IR$^2$L.


\section{Discussion and Conclusion}
The modular network approach is a promising avenue for achieving safe online reinforcement learning. Without a computationally expensive outer meta-learning loop, IR$^2$L  shows that an instinct module can be efficiently trained that can be transferred to multiple tasks and different policies than the ones from a pre-training phase.  
Although the results are encouraging, there is still work to be done in finding instincts with an optimal and predictable trade-off between safety and performance, as well as better guarantees of generality in task transfer.  
The immediate future work involves implementing the IR$^2$L  architecture in CARLA self-driving vehicle \citep{Dosovitskiy17} to protect the car agents during cross-task adaptations.

\section*{Acknowledgments}
This work was supported by the Lifelong Learning Machines program from DARPA/MTO under Contract No. FA8750-18-C-0103. Any opinions, findings, and conclusions, or recommendations expressed in this material are those of the author(s) and do not necessarily reflect the views of DARPA.

\footnotesize
\bibliographystyle{apalike}
\bibliography{bibliography} 

\end{document}